\documentclass[letterpaper,10pt,conference]{ieeeconf}

\usepackage{booktabs}
\usepackage{graphicx}
\usepackage{xcolor}
\usepackage{microtype}
\usepackage{amsmath,amssymb}
\usepackage[nocompress]{cite}
\makeatletter
\let\NAT@parse\undefined%
\makeatother
\usepackage[hidelinks]{hyperref}

\IEEEoverridecommandlockouts%
\title{\LARGE\bfseries
TacSpin: Tactile Tracking of In-Gripper Object Rotation on SO(3) via Axis-Magnitude Estimation}

\author{Yuhao Wu$^{\dagger}$, Linhao Song$^{\dagger}$ and Huan Yin%
\thanks{$^{\dagger}$Yuhao Wu and Linhao Song contributed equally to this work.}%
\thanks{All authors are with the School of AI and Robotics, Hunan University.}
\thanks{Corresponding Author: Huan Yin.}
}

\begin{document}

\maketitle
\thispagestyle{empty}
\pagestyle{empty}
\raggedbottom%

\begin{abstract}
Tracking object motion within a gripper provides essential feedback for robotic manipulation. However, limited contact coverage and uneven deformation make tactile estimation challenging. In this study, we present TacSpin, a method that tracks in-gripper object rotation on SO(3) using only two vision-based tactile sensors. The key idea is to estimate the axis and magnitude of the rotation response separately, rather than fitting a rigid-body model by least squares. Each finger's response is further refined with contact-boundary cues. The two fingers are then fused according to their agreement. On standard objects, TacSpin achieves lower mean checkpoint errors. Qualitative demonstrations on real-world objects, using calibration transferred from a standard sphere, illustrate tracking under different contact geometries.
\end{abstract}
\section{INTRODUCTION}
\label{sec:introduction}

Estimating object motion within a gripper is critical for precise robotic manipulation, such as object reorientation and assembly. During these tasks, an object may rotate or translate relative to the fingers because of object slip or manually applied disturbances. Visual methods can estimate object pose~\cite{wen2024foundationpose}, but the end effector can occlude the object during manipulation~\cite{suresh2024neuralfeels}.

Tactile sensing directly measures the contact surface and provides feedback for robotic perception and manipulation~\cite{lepora2026tactile}. Vision-based tactile sensors (VBTS) capture images of elastomer deformation to measure local contact geometry and deformation~\cite{zhou2026vision}. These sensors provide high-resolution observations~\cite{yuan2017gelsight,lambeta2020digit}, enabling object motion tracking through tactile registration or surface normal alignment~\cite{sodhi2022patchgraph,huang2025normalflow}. However, the small contact region can make motion estimation ambiguous, particularly on textureless surfaces or symmetric objects~\cite{zhang2026tacloc}. Resolving this ambiguity may require additional tactile contacts or external visual observations~\cite{suresh2024neuralfeels,suresh2023midastouch}.

Recently, TacSE3~\cite{liao2026tacse3} demonstrated that object motion relative to the gripper can be estimated from low-texture tactile observations using a rigid body model. It derives planar translation from object motion and estimates rotation by fitting this model to tactile shear responses using least squares. In practice, uneven deformation and local slip can produce different response magnitudes across a contact region. On the other hand, contact constraints imposed by a parallel-jaw gripper can limit changes in the object’s rotation axis. For example, a spoon rotating within a constrained grasp may be approximated as having a single rotational degree of
freedom. From an estimation perspective, a related strategy has been explored in vision-based spin detection for robotic table tennis, where a method based on tracking the ball’s logo estimates the rotation axis before the angular speed~\cite{tebbe2020spin}. These observations and considerations motivate the following question: \textit{Can decoupling axis and magnitude estimation improve tactile rotation tracking?}

In this study, we present TacSpin for tracking in-gripper object rotation. For each finger, we estimate the axis and magnitude of the rotation response using normalized directional voting and the median of local scalar fits, respectively. Each rotation response undergoes boundary circulation refinement, and the two finger rotation responses are fused using mirror alignment and agreement-based attenuation. Offline calibration maps the fused response to rotation increments, which are accumulated on $\mathrm{SO}(3)$ to track rotation relative to the initial grasp. Overall, our contributions are:
\begin{itemize}
    \item We formulate and develop an axis-magnitude estimator, TacSpin, to track the object rotation using the responses of two VTBS.

    \item We evaluate TacSpin quantitatively through baseline comparisons and component ablations, and provide qualitative demonstrations on real-world objects.
\end{itemize}

\begin{figure}[t]
\centering
\includegraphics[width=\columnwidth]{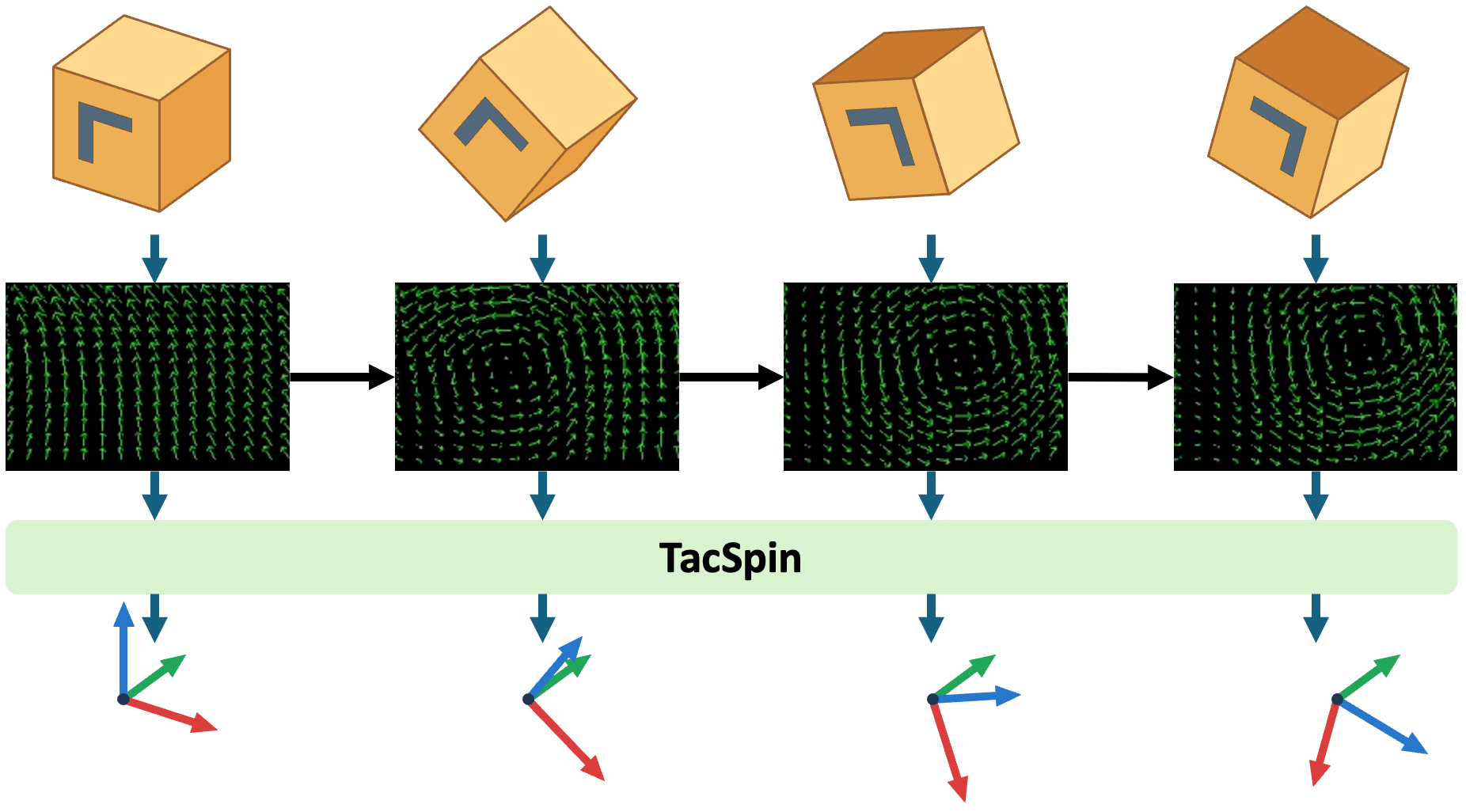}
\caption{Object rotation deforms the VBTS elastomer, thereby producing shear and normal responses. TacSpin estimates the axis and magnitude of a rotation response from these observations and accumulates rotation increments on $\mathrm{SO}(3)$.}%
\label{fig:teaser}
\end{figure}

\section{RELATED WORK}
\label{sec:related_work}

We first review tactile pose estimation of an object from the first touch and then discuss in-hand pose tracking.

\subsection{Object Pose Estimation from Touch}

Initial pose estimation recovers the pose of a stationary or quasi-static object from one or more probing contacts. VBTS observe local surface geometry and deformation~\cite{yuan2017gelsight,lambeta2020digit}. Early approaches used GelSight keypoints with RANSAC, tactile occupancy grids, or covariance features to match observations to object surfaces~\cite{li2014localization,pezzementi2011object,bimbo2016inhand}.

For 6-DoF estimation, Petrovskaya and Khatib combined an analytical contact model with Monte Carlo localization~\cite{petrovskaya2011global}, while Tac2Pose learned object-specific embeddings from rendered contact shapes~\cite{bauza2023tac2pose}. MidasTouch accumulated simulated tactile codes over sliding contact, and Tactile-Filter combined particle filtering with active action selection~\cite{suresh2023midastouch,ota2023tactilefilter}. These methods compare observations with expected contacts synthesized from object models. Inspired by registration methods, TacLoc formulates one-shot estimation as partial-to-full point cloud registration without tactile rendering or object-specific training~\cite{zhang2026tacloc}. YOTO localizes two simultaneous tactile contacts on a known object surface using a coarse-to-fine point cloud network, then recovers pose from surface normals and calibrated sensor poses~\cite{ye2026yoto}.

\subsection{In-Hand Object Pose Tracking}

Once an initial pose is available, in-hand tracking estimates the time-varying object pose or motion between frames relative to the fingers. PatchGraph combines learned surface normals, local registration, and factor graph optimization, whereas NormalFlow aligns successive normal maps for 6-DoF tracking~\cite{sodhi2022patchgraph,huang2025normalflow}. For quasi-static refinement, Shao \textit{et al.}~\cite{shao2026physics} incorporated whole-hand contact geometry and force constraints into a sliding-window particle filter on $\mathrm{SE}(3)$. Limited tactile coverage can cause ambiguity and drift, motivating fusion with global visual cues. Dikhale \textit{et al.}, ViHOPE, and NeuralFeels employ feature fusion, shape completion, and joint optimization of shape and pose, respectively~\cite{dikhale2022visuotactile,li2023vihope,suresh2024neuralfeels}. V-HOP and ViTaSCOPE further address continuous tracking and contact estimation~\cite{li2025vhop,lee2025vitascope}, while HIVE-6D refines visual pose estimates through tactile rendering and comparison and selects modalities according to uncertainty~\cite{liu2026hive6d}. These visuotactile approaches improve spatial coverage but require cross-modal calibration and synchronization and remain dependent on visual availability or learned multimodal models.

Smooth, weakly textured, or locally repetitive contact regions can make image correspondences unreliable and local geometric alignment ambiguous. TacSE3~\cite{liao2026tacse3} addresses this challenge by constructing a 3D response field from normal and shear responses for rigid-body motion estimation on $\mathrm{SE}(3)$. It estimates rotation by least-squares fitting and averages the two finger estimates after mirror alignment. Building on this response-based formulation, TacSpin tracks relative rotation on $\mathrm{SO}(3)$ for objects held in a parallel-jaw gripper. Unlike TacSE3, it first estimates the axis and magnitude of each rotation response separately, and then refines each response with boundary circulation, and fuses the two fingers according to their agreement.
\section{METHODOLOGY}
\label{sec:methodology}

\begin{figure*}[t]
\centering
\includegraphics[width=\textwidth]{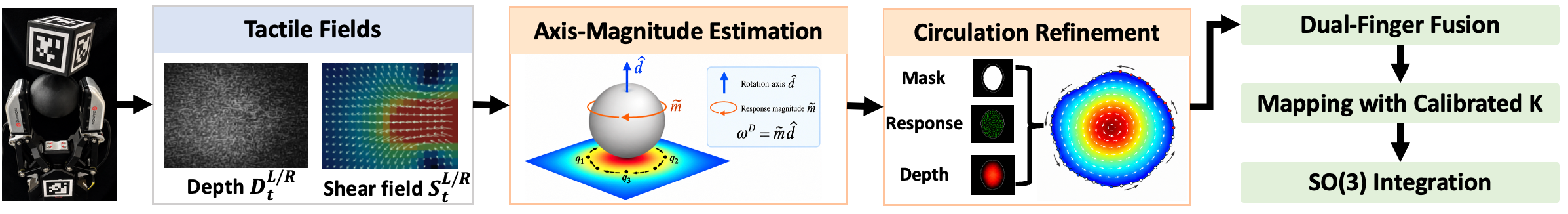}
\caption{Overview of TacSpin. Each finger provides a rotation response through axis-magnitude estimation and boundary circulation refinement. Agreement-based dual-finger fusion and calibration yield rotation increments that are accumulated on $SO(3)$.}%
\label{fig:framework}
\end{figure*}

We consider an object grasped and held between two tactile sensors in a gripper. Given the depth and shear fields from both sensors, TacSpin estimates the object rotation relative to an initial stable grasp. We denote the gripper base frame by $B$ and represent the relative rotation estimate at frame $t$ by $\widehat{\mathbf R}_t\in SO(3)$, where $SO(3)$ denotes the group of three-dimensional rotations and $\widehat{\mathbf R}_0=\mathbf I$.

For finger $j\in\{L,R\}$, let $\mathbf D_t^j$ and $\mathbf S_t^j$ denote the depth field and the two-channel shear field at frame $t$. The normal response $h_t^j$ and shear response $\mathbf v_t^j$ are
\begin{equation}
    h_t^j=\mathbf D_t^j-\mathbf D_0^j,
    \qquad
    \mathbf v_t^j=\mathbf S_t^j-\mathbf S_0^j,
    \label{eq:tactile_responses}
\end{equation}
where $\mathbf D_0^j$ is the initial depth reference and $\mathbf S_0^j$ is the initial shear field recorded after establishing a stable grasp. The shear response $\mathbf v_t^j$ represents deformation-induced displacement relative to this initial shear reference. From the normal response, we extract a contact region $\Omega_t^j$ that excludes weak measurements. Each contact sample pairs a location with its measured shear response. For sample index $i$, the local coordinates are $\mathbf q_{t,i}^j=[x_{t,i}^j,y_{t,i}^j,z_{t,i}^j]^{\mathsf T}$: $x$ and $y$ are coordinates centered in the image and converted to millimeters, and $z$ is the smoothed normal response.

\subsection{Single-Finger Response Estimation}
\label{subsec:single_finger}

As shown in Figure~\ref{fig:framework}, we first estimate the axis and magnitude of each finger's rotation response, then apply boundary circulation refinement. The two refined rotation responses are subsequently fused to track object rotation.

\subsubsection{Axis-Magnitude Estimation}
\begin{figure}[t]
\centering
\includegraphics[width=0.95\columnwidth]{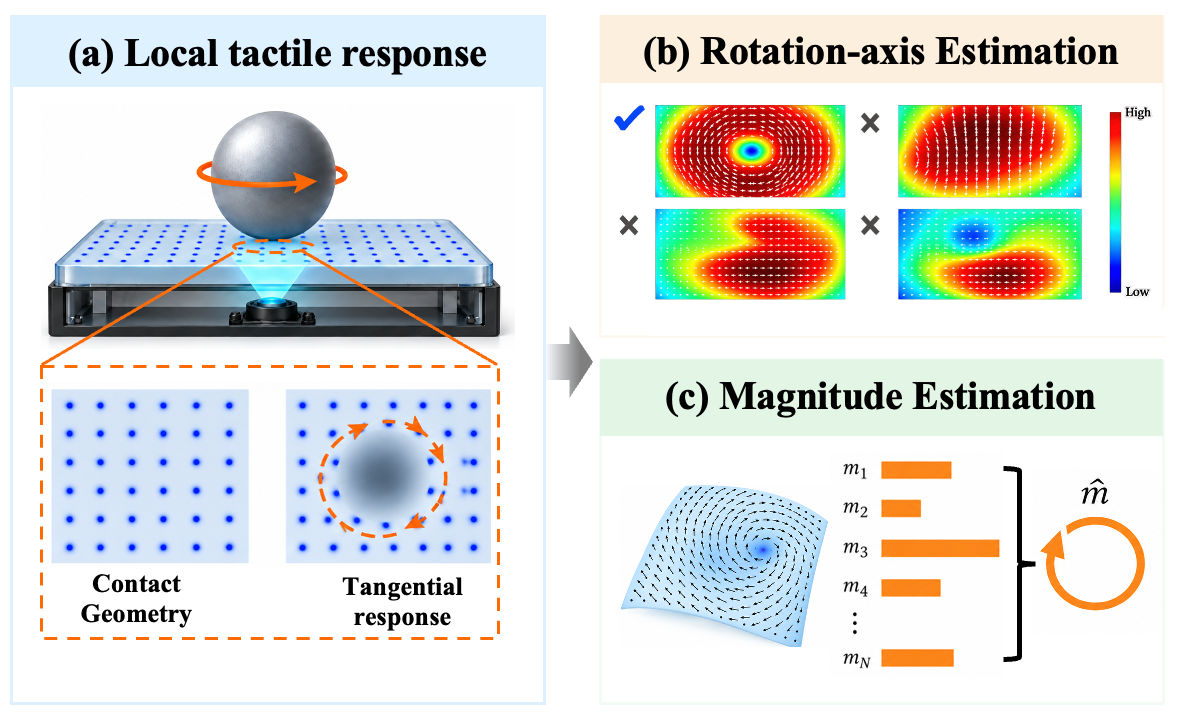}
\caption{Axis-magnitude estimation of a tactile response. Directional voting estimates its axis, and the median of local scalar fits provides its magnitude.}%
\label{fig:dmdr}
\end{figure}

TacSE3~\cite{liao2026tacse3} estimates rotation by fitting a rigid-body model to tactile shear responses using least squares. Large local response deviations can dominate this squared-error fit. We estimate the axis and magnitude of the rotation response separately, as illustrated in Figure~\ref{fig:dmdr}: directional voting selects the axis, and the median of local scalar fits determines the magnitude. Frame and finger indices are omitted below for clarity.

We express each rotation response as $m\mathbf d$. The unit axis $\mathbf d=[d_x,d_y,d_z]^{\mathsf T}$ specifies the direction of this rotation response in the local sensor frame, and $m$ is its magnitude. For a candidate axis $\mathbf d$, the predicted shear response at sample $i$, before scaling by $m$, is
\begin{equation}
    \mathbf u_i(\mathbf d)
    =
    \begin{bmatrix}
        d_yz_i-d_zy_i\\
        d_zx_i-d_xz_i
    \end{bmatrix}.
    \label{eq:theoretical_tangential_response}
\end{equation}
$x_i$ and $y_i$ locate the sample on the sensor plane, and $z_i$ is its smoothed normal response. Equation~\eqref{eq:theoretical_tangential_response} gives the tangential components of $\mathbf d\times\mathbf q_i$; the fitting residual is $m\mathbf u_i(\mathbf d)-\mathbf v_i$. We neglect small translations parallel to the sensor plane.

We select the axis with the highest directional voting score:
\begin{equation}
    \widehat{\mathbf d}
    =
    \underset{\mathbf d\in\mathcal U}{\operatorname{arg\,max}}\,
    \sum_{i=1}^{N}
    \frac{\mathbf u_i(\mathbf d)^{\mathsf T}\mathbf v_i}
    {\|\mathbf u_i(\mathbf d)\|\,\|\mathbf v_i\|},
    \label{eq:dmdr_direction}
\end{equation}
where $\mathcal U$ is a set of unit axes sampled over the sphere, $N$ is the number of valid contact samples, and $\|\cdot\|$ denotes the Euclidean norm. The directional vote of each sample is the cosine similarity between its predicted and measured shear responses, giving each sample equal weight regardless of response magnitude.

After selecting the axis, we fit a scalar coefficient between the predicted and measured shear responses at each sample and take the median across the contact region. This yields the initial rotation response $\widetilde{\boldsymbol{\omega}}^{D}$, where the tilde denotes an uncalibrated response and superscript $D$ denotes the initial axis-magnitude estimate:
\begin{equation}
\widehat m =
\underset{i}{\operatorname{median}}
\left(
\frac{\mathbf v_i^{\mathsf T}\mathbf u_i(\widehat{\mathbf d})}
{\|\mathbf u_i(\widehat{\mathbf d})\|^2}
\right),
\quad
\widetilde{\boldsymbol{\omega}}^{D} = \widehat m\,\widehat{\mathbf d}.
\label{eq:dmdr_rotation_estimate}
\end{equation}
The axis and magnitude are thus estimated sequentially: voting first selects the axis $\widehat{\mathbf{d}}$, and the magnitude $\widehat{m}$ is then obtained as the median of local scalar fits along that axis.

\subsubsection{Boundary Circulation Refinement}
\begin{figure}[t]
\centering
\includegraphics[width=0.95\columnwidth]{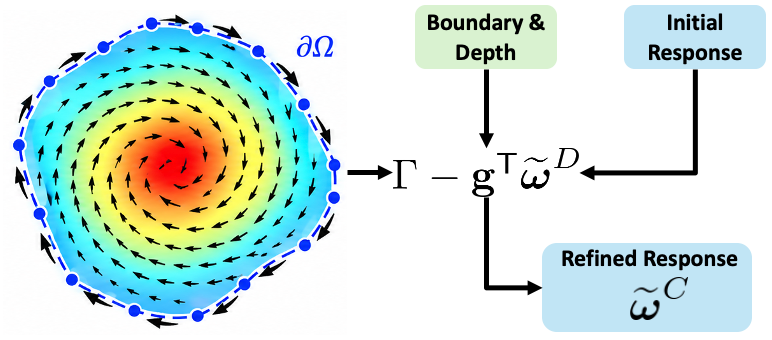}
\caption{Boundary circulation refinement of a tactile rotation response. The difference between measured and predicted boundary circulation guides a scaled correction of the initial response.}
\label{fig:circ}
\end{figure}

Local slip and uneven contact deformation can disturb individual shear responses. To address this, we use the shear pattern along the contact boundary to check and refine the initial estimate, as illustrated in Figure~\ref{fig:circ}. For example, rotation about the sensor normal tends to produce a clockwise or counterclockwise shear pattern around the contact. Circulation measures this overall tendency.

To compute circulation, we traverse the closed contact boundary $\partial\Omega$ in a fixed direction. At each location, we project the measured shear response onto the local boundary tangent and accumulate this signed contribution along the boundary. Contributions following the traversal direction are positive, while opposing contributions are negative. Their total gives the measured circulation:
\begin{equation}
    \Gamma
    =
    \oint_{\partial\Omega}
    \left(v_x\,\mathrm dx+v_y\,\mathrm dy\right),
    \label{eq:circulation}
\end{equation}
where $v_x$ and $v_y$ are the measured shear response components, and $\mathrm dx$ and $\mathrm dy$ are coordinate increments along the boundary. A spatially uniform additive shear component cancels in the integral over a fixed closed boundary.

Substituting the magnitude-scaled shear response predicted by Equation~\eqref{eq:theoretical_tangential_response} into the circulation integral gives the prediction $\mathbf g^{\mathsf T}\widetilde{\boldsymbol{\omega}}^{D}$. The vector $\mathbf g$ collects the contribution of each rotation response component and depends on the boundary coordinates and normal response:
\begin{equation}
    \mathbf g
    =
    \oint_{\partial\Omega}
    \begin{bmatrix}
        -z\,\mathrm dy\\
        z\,\mathrm dx\\
        x\,\mathrm dy-y\,\mathrm dx
    \end{bmatrix}.
    \label{eq:circulation_constraint}
\end{equation}
Each component of $\mathbf g$ describes how a rotation response about the corresponding sensor axis contributes to circulation. When the contact boundary lies in a plane parallel to the sensor surface, its circulation constrains only the rotation component about the sensor normal. For a parallel-jaw grasp, however, this is precisely the relevant component: an object held between two opposing fingertips most easily spins about the axis through the two contacts, which is normal to the sensor surface.

We compare the measured circulation $\Gamma$ with the prediction $\mathbf g^{\mathsf T}\widetilde{\boldsymbol{\omega}}^{D}$. Their difference, $\Delta\Gamma=\Gamma-\mathbf g^{\mathsf T}\widetilde{\boldsymbol{\omega}}^{D}$, specifies the correction needed along $\mathbf g$. The smallest correction that matches the measured circulation is parallel to $\mathbf g$, leaving perpendicular components unchanged. The boundary measurement therefore adjusts the component it constrains without replacing the full three-dimensional estimate.

We scale this correction according to the signed boundary contributions, reducing it when positive and negative contributions cancel. Denoting the resulting correction by $\delta\boldsymbol{\omega}$, we obtain the refined response:
\begin{equation}
\widetilde{\boldsymbol{\omega}}^{C}
=
\widetilde{\boldsymbol{\omega}}^{D}
+
\delta\boldsymbol{\omega},
\label{eq:circulation_refinement}
\end{equation}
where superscript $C$ denotes circulation refinement.

\subsection{Agreement-Based Dual-Finger Fusion}
\label{subsec:dual_finger}

The two VBTS observe the same object rotation through different contact regions. Their rotation responses can differ because of unequal deformation or local disturbances, even after refinement for each finger. After mirror alignment, dual-finger fusion uses this disagreement for agreement-based attenuation of individual components of the common response.

We first express both rotation responses in the left sensor frame, denoted by $S$. For the opposing sensor arrangement, mirror alignment from the right to the left sensor frame uses the diagonal matrix $\mathbf M=\operatorname{diag}(1,-1,-1)$, giving $\overline{\boldsymbol{\omega}}_t^R=\mathbf M\widetilde{\boldsymbol{\omega}}_t^{R,C}$. The overbar denotes the mirror-aligned right rotation response, which has undergone circulation refinement and remains uncalibrated. Their common ($+$) and differential ($-$) responses are
\begin{equation}
    \widetilde{\boldsymbol{\omega}}_t^{\pm}
    =\tfrac12\left(
        \widetilde{\boldsymbol{\omega}}_t^{L,C}
        \pm\overline{\boldsymbol{\omega}}_t^R\right).
\label{eq:common_differential_components}
\end{equation}
in which $\widetilde{\boldsymbol{\omega}}_t^+$ encodes the aligned averaging response, while the differential response $\widetilde{\boldsymbol{\omega}}_t^-$ measures disagreement between the two rotation responses. For axis $a\in\{x,y,z\}$, subscript $a$ selects the scalar component. To attenuate components with large relative disagreement, we use an exponential factor with the same functional form as the Welsch robust weight~\cite{babin2019analysis}. The fused component $\widetilde{\omega}_{t,a}^S$ in frame $S$ is defined as
\begin{equation}
    \widetilde{\omega}_{t,a}^S
    =
    \widetilde{\omega}_{t,a}^+
    \exp\!\left[
        -\gamma_a
        \left(\frac{\widetilde{\omega}_{t,a}^-}{\widetilde{\omega}_{t,a}^+}\right)^2
    \right],
    \label{eq:mcdf_fusion}
\end{equation}
where $\gamma_a\geq0$ controls attenuation along axis $a$ and $\exp$ is the scalar exponential function. The fused component is defined as zero when $\widetilde{\omega}_{t,a}^+=0$. The exponential factor equals one when the fingers agree and decreases with relative disagreement for $\gamma_a>0$, regardless of the sign of that disagreement. Applying different factors to different components can change both the axis and magnitude of the fused rotation response before calibration and integration.

\subsection{Calibration and Rotation Tracking}
\label{subsec:calibration_tracking}

The estimated magnitude reflects tactile response strength and does not directly represent a rotation angle or angular speed. We therefore calibrate a model for discrete rotation updates using a full matrix $\mathbf K\in\mathbb R^{3\times3}$. For the fused response $\widetilde{\boldsymbol{\omega}}_t^S$, the calibrated rotation increment is
\begin{equation}
    \delta\widehat{\boldsymbol{\theta}}_t
    =
    \mathbf R_{BS}\mathbf K\widetilde{\boldsymbol{\omega}}_t^S\Delta t_t,
    \label{eq:full_k_calibration}
\end{equation}
where $\mathbf R_{BS}$ is the fixed left-sensor-to-base rotation, $\Delta t_t$ is the elapsed time in seconds between frames $t-1$ and $t$, and $\delta\widehat{\boldsymbol{\theta}}_t$ is expressed in radians. The fitted matrix $\mathbf K$ captures response scale and cross-axis coupling in this mapping.

At runtime, we accumulate these increments on $SO(3)$ through right multiplication using the exponential map~\cite{sola2018micro}:
\begin{equation}
    \widehat{\mathbf R}_t
    =
    \widehat{\mathbf R}_{t-1}
    \operatorname{Exp}\!\left(
        [\delta\widehat{\boldsymbol{\theta}}_t]_\times
    \right),
    \label{eq:so3_integration}
\end{equation}
where $[\cdot]_\times$ maps a three-dimensional vector to its skew-symmetric matrix and $\operatorname{Exp}$ is the matrix exponential. The increment parameterizes the rotation composed on the right of the current estimate. We fit $\mathbf K$ offline using this same update rule and reference rotation trajectories, as described in Section~\ref{subsec:calibration_evaluation}. The output $\widehat{\mathbf R}_t$ tracks rotation relative to the initial grasp; when a scalar angle is needed, we use $\|\operatorname{Log}(\widehat{\mathbf R}_t)^\vee\|$, where $\operatorname{Log}$ is the rotation matrix logarithm and $\vee$ maps its skew-symmetric result back to a three-dimensional vector.

\section{Experiments}\label{sec:experiment}

We evaluate rotation tracking on sequences of standard objects and use component ablations to examine how the results depend on the motion and error metric.

\subsection{Experimental Setup}

\subsubsection{Platform and Metric}

We build an experimental platform for tactile rotation tracking, as shown in Figure~\ref{fig:experimental_settings}. A fixed parallel-jaw gripper holds the object between two opposing vision-based tactile sensors, while an external camera observes an AprilTag cube attached to the grasped object as a tracking fixture and a reference tag on the gripper. The objects are rotated manually while the gripper remains fixed. Camera measurements are used only for offline calibration and evaluation, and online rotation estimation relies only on the two tactile streams.

For evaluation, let $O$ and $C$ denote the object and camera frames, respectively. The poses $\mathbf T_C^O(t)$ and $\mathbf T_C^B(t)$ map coordinates from the object and gripper base frames into the camera frame at time $t$. The reference object pose relative to the gripper is
\begin{equation}
    \mathbf T_B^O(t)
    = \left(\mathbf T_C^B(t)\right)^{-1}\mathbf T_C^O(t).
    \label{eq:apriltag_object_pose}
\end{equation}
Since TacSpin estimates rotation only, we take the rotation block of $\mathbf{T}_B^O(t)$ as the reference rotation $\mathbf{Q}_{\mathrm{ref}}(t)$ and discard the translation. The estimated rotation $\mathbf{Q}_{\mathrm{est}}(t)$ is the output $\widehat{\mathbf{R}}_t$ of Equation~(11). Because $\widehat{\mathbf{R}}_t$ is a relative rotation accumulated from the start of the sequence, both trajectories are expressed relative to their first valid paired timestamp $t_0$:
\begin{equation}
    \begin{aligned}
        \overline{\mathbf R}_{\mathrm{ref}}(t)
        &= \mathbf Q_{\mathrm{ref}}(t_0)^{\mathsf T}\mathbf Q_{\mathrm{ref}}(t),\\
        \overline{\mathbf R}_{\mathrm{est}}(t)
        &= \mathbf Q_{\mathrm{est}}(t_0)^{\mathsf T}\mathbf Q_{\mathrm{est}}(t).
    \end{aligned}
    \label{eq:relative_ground_truth}
\end{equation}
The resulting relative rotations are denoted by $\overline{\mathbf R}_{\mathrm{ref}}(t)$ and $\overline{\mathbf R}_{\mathrm{est}}(t)$. Their rotation error is the geodesic distance
\begin{equation}
    e_R(t)
    = \left\|
        \operatorname{Log}\!\left(
            \overline{\mathbf R}_{\mathrm{est}}(t)
            \overline{\mathbf R}_{\mathrm{ref}}(t)^{\mathsf T}
        \right)^{\vee}
    \right\|.
    \label{eq:rotation_error}
\end{equation}
We report the endpoint error $e_R(t_{\mathrm{end}})$ and the mean trajectory error $E_{\mathrm{traj}}=N_f^{-1}\sum_{k=1}^{N_f}e_R(t_k)$, where $N_f$ counts valid evaluation frames, $t_k$ is the timestamp of evaluation frame $k$, and $t_{\mathrm{end}}=t_{N_f}$ is the final valid timestamp. All errors are expressed in degrees in this section.

\subsubsection{Calibration and Evaluation}
\label{subsec:calibration_evaluation}

We calibrate the nine entries of $\mathbf K$ offline using synchronized tactile responses and AprilTag orientations. Starting from a linear initialization, we refine $\mathbf K$ by integrating the tactile predictions and minimizing mean trajectory and endpoint errors with regularization.

\begin{figure}[t]
    \centering
    \includegraphics[width=0.95\columnwidth]{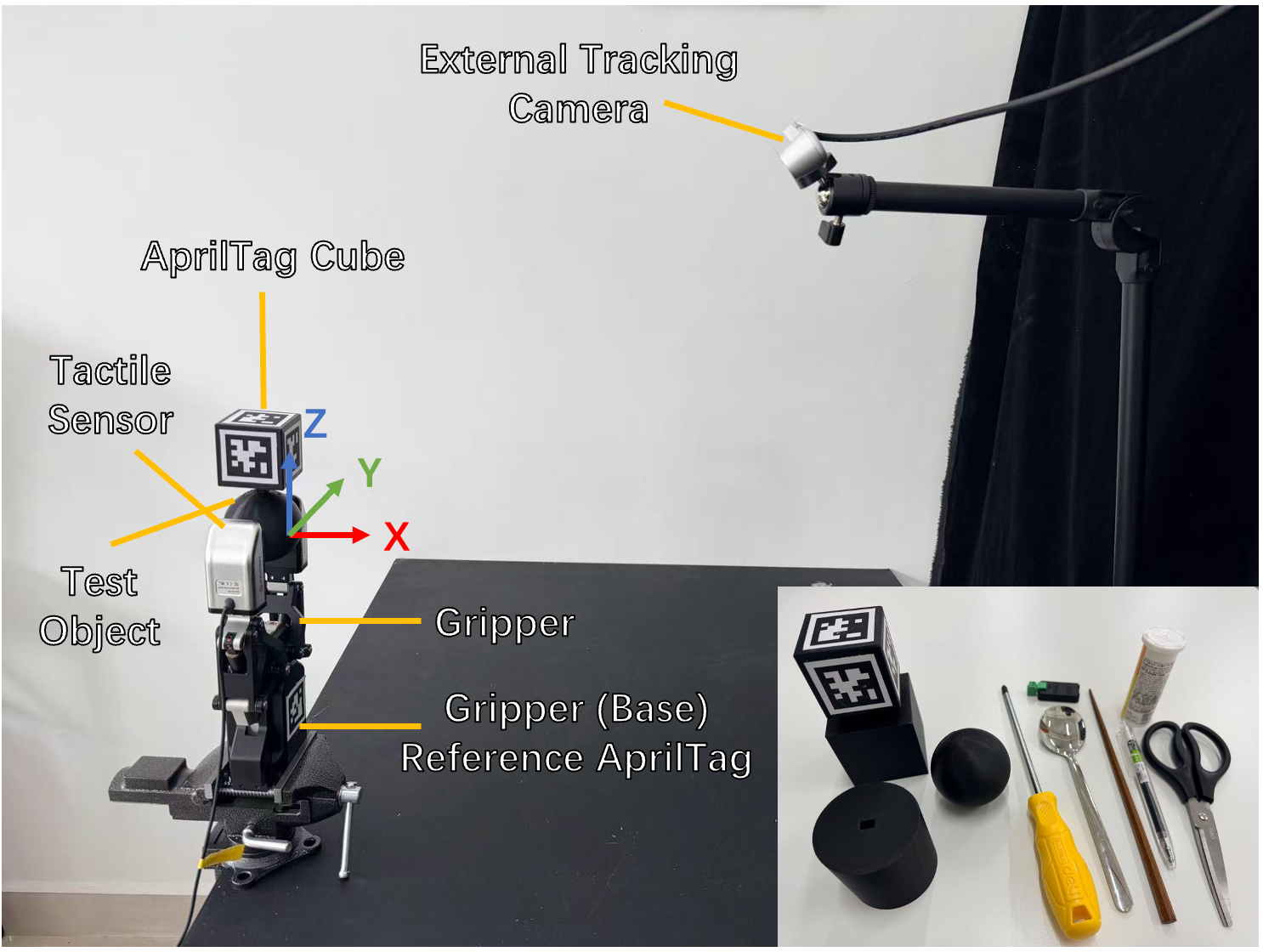}
    \caption{Experimental setup. An external camera observes the AprilTag cube attached to the object and the reference tag on the gripper. The inset at the lower right shows the standard objects used for quantitative evaluation and some of the real-world objects used for qualitative demonstrations.}
    \label{fig:experimental_settings}
\end{figure}

Three standard objects (a cube, a cylinder, and a sphere) were 3D printed for quantitative evaluation. We collect synchronized tactile and AprilTag data for a cube rotating about its initial $Y$ axis, a cylinder about $X$ and $Y$, and a sphere undergoing mixed-axis motion with a changing rotation axis. The axis directions are shown in Figure~\ref{fig:experimental_settings}. Fixed-axis sequences are evaluated at $30^\circ$, $60^\circ$, and $90^\circ$ reference-angle checkpoints within each sequence. The checkpoint error is $e_R(t)$ at the final valid paired frame of the evaluation segment associated with each checkpoint. The sphere is evaluated over the complete sequence. We also test TacSpin on the real-world objects, and some of them are shown in Figure~\ref{fig:experimental_settings} and~\ref{fig:real_object_tracking}. These real-world object sequences have no external pose reference and serve as qualitative demonstrations. We use the same $\mathbf K$ calibrated on the standard sphere for these objects to qualitatively examine whether the calibration transfers across object geometries.

Sequences of the standard objects are indexed by repetition number. For each fixed-axis condition, repetitions 01--03 calibrate $\mathbf K$ and 04--08 form the test set, giving three calibration and five test sequences per condition. For the sphere, repetitions 01--05 provide five calibration sequences and 06--08 provide three test sequences. In total, the dataset of standard objects contains 14 calibration and 18 test sequences. For each object, a single $\mathbf{K}$ is fitted jointly across axes, and held fixed during testing.

\subsubsection{Implementation and Baseline}

We compare TacSpin with our implementation of the response-based state estimator in TacSE3~\cite{liao2026tacse3}. This baseline uses three diagonal gains, applied componentwise to the TacSE3 rotation response. We also test TacSE3 + full-K, which retains the least-squares-based estimator and replaces the diagonal gains with a full $3\times3$ matrix. For each standard object, all methods use the same calibration/test split and calibration loss, with parameters fitted only on the calibration sequences and kept fixed during testing.

\subsection{Tracking Evaluation on Standard Objects}

Table~\ref{tab:standard_results} reports mean $\pm$ sample standard deviation across test sequences for each fixed-axis condition and for sphere motion. The fixed-axis mean averages the condition means. Sphere mean trajectory and endpoint errors are averaged with equal sequence weights.

\begin{table}[t]
    \caption{Rotation errors on standard objects ($^\circ$).}%
    \label{tab:standard_results}
    \centering
    \small
    \setlength{\tabcolsep}{3pt}
    \renewcommand{\arraystretch}{1.06}
    \begin{tabular*}{\columnwidth}{@{\extracolsep{\fill}}lccc@{}}
        \toprule
        Condition & TacSE3 & \shortstack{TacSE3\\+ full-K} & TacSpin \\
        \midrule
        \multicolumn{4}{l}{\textit{(a) Fixed-axis checkpoint error (5 test sequences per condition)}} \\
        Cube $Y$, $30^\circ$ & $6.5\pm5.1$ & $11.3\pm7.7$ & $\mathbf{5.5}\pm2.8$ \\
        Cube $Y$, $60^\circ$ & $17.9\pm8.1$ & $29.1\pm14.4$ & $\mathbf{11.9}\pm7.7$ \\
        Cube $Y$, $90^\circ$ & $25.7\pm8.2$ & $43.1\pm14.8$ & $\mathbf{15.3}\pm10.5$ \\
        \addlinespace[4pt]
        Cylinder $X$, $30^\circ$ & $10.2\pm4.0$ & $8.0\pm3.8$ & $\mathbf{6.8}\pm3.0$ \\
        Cylinder $X$, $60^\circ$ & $19.9\pm3.8$ & $17.6\pm6.6$ & $\mathbf{14.1}\pm9.2$ \\
        Cylinder $X$, $90^\circ$ & $29.1\pm6.2$ & $27.8\pm12.2$ & $\mathbf{22.4}\pm11.9$ \\
        \addlinespace[4pt]
        Cylinder $Y$, $30^\circ$ & $7.2\pm2.4$ & $7.1\pm2.2$ & $\mathbf{5.4}\pm3.2$ \\
        Cylinder $Y$, $60^\circ$ & $13.5\pm5.2$ & $14.6\pm5.1$ & $\mathbf{7.0}\pm4.4$ \\
        Cylinder $Y$, $90^\circ$ & $16.7\pm12.4$ & $20.5\pm11.4$ & $\mathbf{11.4}\pm8.0$ \\
        \midrule
        Fixed-axis mean & $16.3$ & $19.9$ & $\mathbf{11.1}$ \\
        \midrule
        \multicolumn{4}{l}{\textit{(b) Sphere mixed-axis motion (3 test sequences)}} \\
        $E_{\mathrm{traj}}$ & $19.3\pm6.7$ & $\mathbf{17.3}\pm5.9$ & $\mathbf{17.3}\pm5.1$ \\
        $e_R(t_{\mathrm{end}})$ & $\mathbf{24.8}\pm7.9$ & $27.0\pm6.9$ & $34.9\pm27.4$ \\
        \bottomrule
    \end{tabular*}
\end{table}

\begin{figure*}[t]
    \centering
    \includegraphics[width=0.95\textwidth]{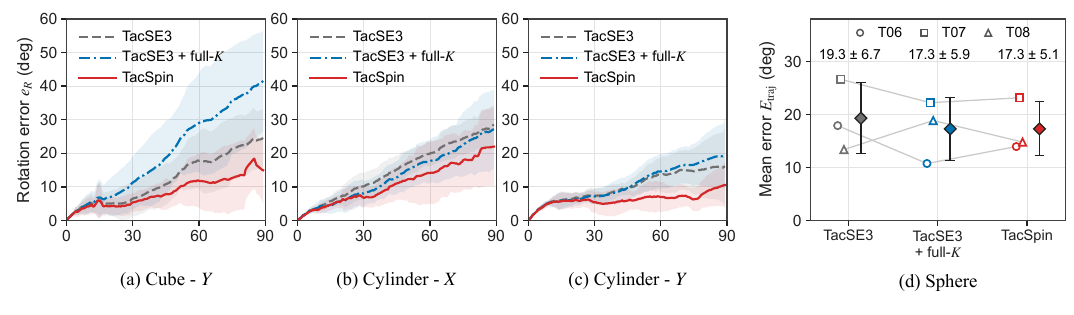}
    \caption{Rotation tracking errors on the standard objects. (a)--(c) Rotation error for the cube about $Y$ and the cylinder about $X$ and $Y$, respectively. The horizontal axis is the reference rotation angle relative to the initial orientation, in degrees. Curves show the mean of five test sequences aligned by reference angle over their common angular range; shaded bands indicate $\pm$ one sample standard deviation. (d) Mixed-axis sphere motion. Open markers show the mean trajectory error of each test sequence (T06--T08 denote test repetitions 06--08), with gray lines connecting the same sequence across methods. Filled diamonds and error bars show the mean and $\pm$ one sample standard deviation across the three sequence-level mean trajectory errors. All plotted sequences are excluded from parameter fitting.}
    \label{fig:standard_object_tracking}
\end{figure*}

TacSpin achieves lower mean checkpoint error than both baselines in all evaluated fixed-axis conditions. To assess consistency across test sequences, we also average checkpoint errors within each sequence. By this measure, TacSpin outperforms TacSE3 on 11 of 15 fixed-axis test sequences and TacSE3 + full-K on 13 of 15. Notably, TacSE3 + full-$\mathbf{K}$ performs worse than TacSE3 on the cube, even though the diagonal $\mathbf{K}$ is a special case of the full matrix. With only three calibration sequences for the cube, the additional off-diagonal parameters appear to overfit. On the cylinder, where six calibration sequences are available, the full $\mathbf{K}$ helps about the $X$ axis but not about the $Y$ axis, suggesting that more calibration data alleviates but does not remove the overfitting. In contrast, the robust per-finger estimates of TacSpin yield a more consistent response under the same parameterization. Figure~\ref{fig:standard_object_tracking}(a)--(c) shows rotation errors $e_R$ against the reference rotation angle.

For mixed-axis sphere motion, TacSE3 + full-K reduces the mean trajectory error relative to TacSE3 and performs comparably to TacSpin. The off-diagonal terms of the full matrix may account for cross-axis coupling in these sequences. Figure~\ref{fig:standard_object_tracking}(d) shows that neither TacSpin nor TacSE3 + full-K consistently outperforms the other across the test sequences. TacSpin has a higher mean endpoint error than both baselines. This partly reflects late-stage error growth in one test sequence, which reaches $64.4^\circ$ at the endpoint despite a mean trajectory error of $23.1^\circ$.

\subsection{Ablation Study}

We compare the full pipeline with variants that remove circulation refinement or the combined mirror alignment and agreement-based attenuation procedure. For each ablated variant, we refit $\mathbf K$ on the same calibration set. More specifically, the variant labeled ``w/o fusion'' removes mirror alignment and agreement-based attenuation. It directly averages the two finger rotation responses without mapping the right response into the left sensor frame or applying agreement-based attenuation. Without mirror alignment, corresponding response components can have opposite signs and cancel in the average. TacSE3 itself averages after mirror alignment, so this ablation measures the combined effect of removing mirror alignment and agreement-based attenuation rather than the effect of attenuation alone.

Table~\ref{tab:component_ablation_heldout} reports errors for each object. For fixed-axis motion, we first average checkpoint errors within each sequence and then average over test sequences for each object, including both cylinder axes. For the sphere, both $E_{\mathrm{traj}}$ and $e_R(t_{\mathrm{end}})$ are averaged over the same test sequences used in Table~\ref{tab:standard_results}.

\begin{table}[t]
  \centering
  \caption{Rotation errors of component ablations ($^\circ$).}%
  \label{tab:component_ablation_heldout}
  \small
  \renewcommand{\arraystretch}{1.08}
  \setlength{\tabcolsep}{2pt}
  \begin{tabular*}{\columnwidth}{@{\extracolsep{\fill}}lcccc@{}}
    \toprule
    & \multicolumn{2}{c}{Mean checkpoint error} & \multicolumn{2}{c}{Sphere} \\
    \cmidrule(lr){2-3}\cmidrule(l){4-5}
    Variant & Cube & Cylinder & $E_{\mathrm{traj}}$ & $e_R(t_{\mathrm{end}})$ \\
    \midrule
    w/o circ. refinement & 17.0 & 12.5 & \textbf{16.5} & 30.5 \\
    \mbox{w/o fusion} & 46.5 & 55.9 & 22.4 & \textbf{24.8} \\
    Full TacSpin & \textbf{10.9} & \textbf{11.2} & 17.3 & 34.9 \\
    \bottomrule
  \end{tabular*}
\end{table}

Removing mirror alignment and agreement-based attenuation increases the cube and cylinder checkpoint errors. For the sphere, it increases $E_{\mathrm{traj}}$ despite reducing $e_R(t_{\mathrm{end}})$. Lower endpoint error does not imply lower error throughout the sequence. The benefit of circulation refinement is evident for the cube and cylinder. On the sphere test sequences, removing circulation refinement yields lower mean errors under both metrics. Both components therefore contribute to the fixed-axis performance reported in Table~I, with fusion accounting for the larger share. Their benefit does not carry over to mixed-axis sphere motion.

\subsection{Tracking on Real-World Objects}

\begin{figure*}[t]
    \centering
    \includegraphics[width=0.95\textwidth]{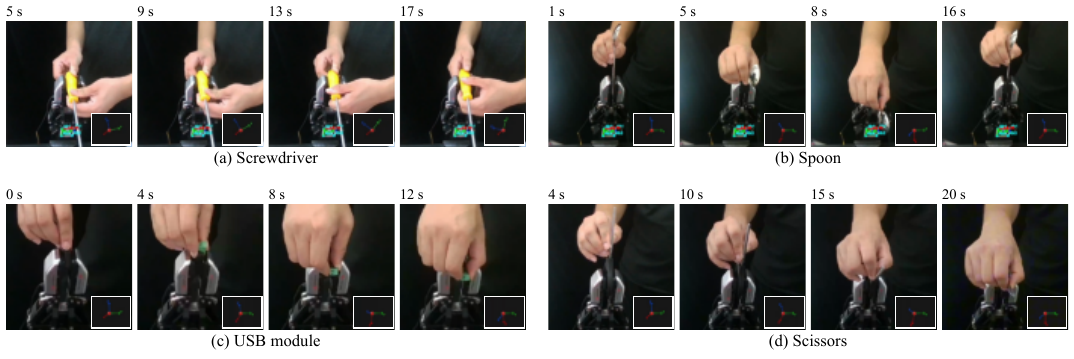}
    \caption{Qualitative rotation tracking of different objects. Insets show the corresponding tactile orientation estimates, with the $x$, $y$, and $z$ axes in red, green, and blue, respectively.}
    \label{fig:real_object_tracking}
\end{figure*}

Figure~\ref{fig:real_object_tracking} shows tactile rotation estimates during manually applied rotations of the real-world objects. Because there is no external pose reference, these demonstrations are excluded from error statistics. It is worth noting that no object-specific calibration or parameter tuning is performed: the same $\mathbf K$ calibrated on the standard sphere is applied directly to all real-world objects, which differ substantially from the calibration object in size, curvature, and surface stiffness. \footnote{Real-world object results are available in the supplementary video.}

\section{CONCLUSION}\label{sec:conclusion}

In this study, we develop TacSpin to track in-gripper object rotation. TacSpin estimates the axis and magnitude of each rotation response separately. Experiments show lower mean errors in the tested fixed-axis conditions. Demonstrations on real-world objects illustrate tracking under different contact geometries.

Based on the experimental results, we also summarize the limitations and future work as follows:
\begin{itemize}
    \item \textit{Mixed-axis motion.}
    The current method shows benefits in the tested fixed-axis conditions, while mixed-axis sphere tracking remains challenging. Future work will improve estimation for motion with changing rotation axes.

    \item \textit{Translation estimation.}
    The current formulation estimates rotation only. We will extend the method to jointly estimate translation and rotation for more general in-gripper motion tracking.

    \item \textit{Downstream manipulation.}
    The current evaluation focuses on tracking performance without closed-loop task execution. We will integrate TacSpin into manipulation controllers and evaluate its utility for tasks such as object reorientation and assembly.
\end{itemize}

\section*{Acknowledgment}
Individual illustrative elements in Figs. 1–4 were generated with ChatGPT. All figure composition, layout, and editing were done by the authors.

\bibliographystyle{IEEEtran}
\bibliography{references}

\end{document}